\documentclass[10pt,twocolumn,letterpaper]{article}

\usepackage{cvpr}
\usepackage{times}
\usepackage{epsfig}
\usepackage{graphicx}
\usepackage{amsmath}
\usepackage{amssymb}

\usepackage{sg-macros}
\usepackage{booktabs}
\usepackage{tabularx}
\usepackage{makecell}

\usepackage[pagebackref=true,breaklinks=true,letterpaper=true,colorlinks,bookmarks=false]{hyperref}

\cvprfinalcopy 

\def\cvprPaperID{1163} 

\ifcvprfinal\fi
\begin{document}

\title{Bottom-Up and Top-Down Attention for Image Captioning\\and Visual Question Answering}

\author{Peter Anderson\textsuperscript{1}\thanks{Work performed while interning at Microsoft.}\hspace{20pt} Xiaodong He\textsuperscript{2}\hspace{20pt} Chris Buehler\textsuperscript{3}\hspace{20pt} Damien Teney\textsuperscript{4}\\ Mark Johnson\textsuperscript{5}\hspace{20pt} Stephen Gould\textsuperscript{1}\hspace{20pt} Lei Zhang\textsuperscript{3}\\
	\normalsize{
		\textsuperscript{1}Australian National University \space
		\textsuperscript{2}JD AI Research \space
		\textsuperscript{3}Microsoft Research \space
		\textsuperscript{4}University of Adelaide \space
		\textsuperscript{5}Macquarie University
	}\\
	\tt\small\textsuperscript{1}firstname.lastname@anu.edu.au,
	\tt\small\textsuperscript{2}xiaodong.he@jd.com,
	\tt\small\textsuperscript{3}\{chris.buehler,leizhang\}@microsoft.com\\
	\tt\small\textsuperscript{4}damien.teney@adelaide.edu.au,
	\tt\small\textsuperscript{5}mark.johnson@mq.edu.au
}
\maketitle

\begin{abstract}
	
Top-down visual attention mechanisms have been used extensively in image captioning and visual question answering (VQA) to enable deeper image understanding through fine-grained analysis and even multiple steps of reasoning. In this work, we propose a combined bottom-up and top-down attention mechanism that enables attention to be calculated at the level of objects and other salient image regions. This is the natural basis for attention to be considered. Within our approach, the bottom-up mechanism (based on Faster R-CNN) proposes image regions, each with an associated feature vector, while the top-down mechanism determines feature weightings. Applying this approach to image captioning, our results on the MSCOCO test server establish a new state-of-the-art for the task, achieving CIDEr / SPICE / BLEU-4 scores of 117.9, 21.5 and 36.9, respectively. Demonstrating the broad applicability of the method, applying the same approach to VQA we obtain first place in the 2017 VQA Challenge.
\end{abstract}

\section{Introduction}

Problems combining image and language understanding such as image captioning~\cite{Chen2015} and visual question answering (VQA)~\cite{balanced_vqa_v2} continue to inspire considerable research at the boundary of computer vision and natural language processing. In both these tasks it is often necessary to perform some fine-grained visual processing, or even multiple steps of reasoning to generate high quality outputs. As a result, visual attention mechanisms have been widely adopted in both image captioning~\cite{scst2016,sentinel,reviewnet,Xu2015} and VQA~\cite{fukui2016multimodal,coatt,askattend,stacked,visual7w}. These mechanisms improve performance by learning to focus on the regions of the image that are salient and are currently based on deep neural network architectures.

In the human visual system, attention can be focused volitionally by top-down signals determined by the current task (e.g., looking for something), and automatically by bottom-up signals associated with unexpected, novel or salient stimuli~\cite{buschman2007top,corbetta2002control}. In this paper we adopt similar terminology and refer to attention mechanisms driven by non-visual or task-specific context as `top-down', and purely visual feed-forward attention mechanisms as `bottom-up'.

\begin{figure}[t]
	\begin{minipage}{.5\linewidth}
		\begin{center}
			\includegraphics[width=1.0\linewidth]{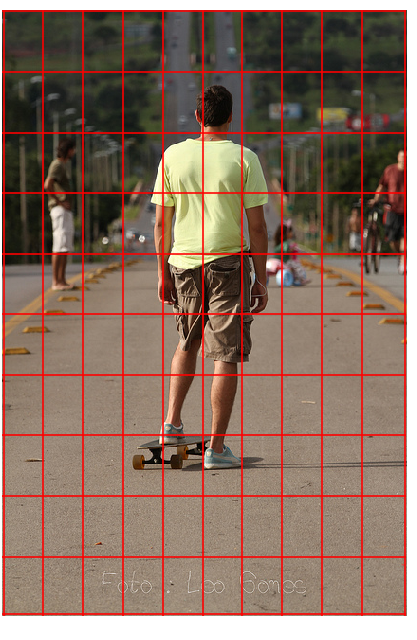}
		\end{center}
	\end{minipage}%
	\begin{minipage}{.5\linewidth}
		\begin{center}
			\includegraphics[width=1.0\linewidth]{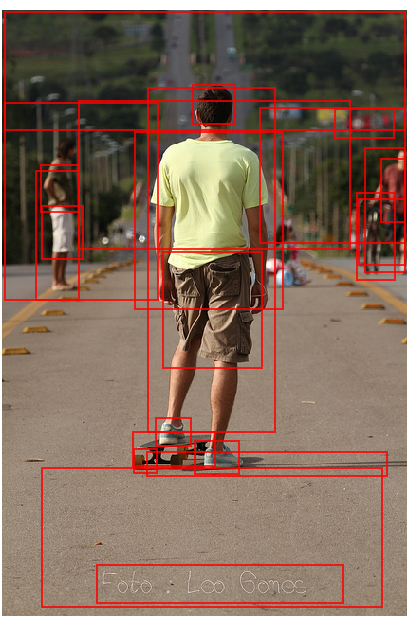}
		\end{center}
	\end{minipage}%
	\caption{Typically, attention models operate on CNN features corresponding to a uniform grid of equally-sized image regions (left). Our approach enables attention to be calculated at the level of objects and other salient image regions (right).}
	\label{fig:concept}
\end{figure}

Most conventional visual attention mechanisms used in image captioning and VQA are of the top-down variety. Taking as context a representation of a partially-completed caption output, or a question relating to the image, these mechanisms are typically trained to selectively attend to the output of one or more layers of a convolutional neural net (CNN). However, this approach gives little consideration to how the image regions that are subject to attention are determined. As illustrated conceptually in \figref{fig:concept}, the resulting input regions correspond to a uniform grid of equally sized and shaped neural receptive fields -- irrespective of the content of the image. To generate more human-like captions and question answers, objects and other salient image regions are a much more natural basis for attention~\cite{egly1994shifting,scholl2001objects}.

In this paper we propose a combined bottom-up and top-down visual attention mechanism. The bottom-up mechanism proposes a set of salient image regions, with each region represented by a pooled convolutional feature vector. Practically, we implement bottom-up attention using Faster R-CNN~\cite{faster_rcnn}, which represents a natural expression of a bottom-up attention mechanism. The top-down mechanism uses task-specific context to predict an attention distribution over the image regions. The attended feature vector is then computed as a weighted average of image features over all regions. 

We evaluate the impact of combining bottom-up and top-down attention on two tasks. We first present an image captioning model that takes multiple glimpses of salient image regions during caption generation. Empirically, we find that the inclusion of bottom-up attention has a significant positive benefit for image captioning. Our results on the MSCOCO test server establish a new state-of-the-art for the task, achieving CIDEr / SPICE / BLEU-4 scores of 117.9, 21.5 and 36.9. respectively (outperforming all published and unpublished work at the time). Demonstrating the broad applicability of the method, we additionally present a VQA model using the same bottom-up attention features. Using this model we obtain first place in the 2017 VQA Challenge, achieving 70.3\% overall accuracy on the VQA v2.0 test-standard server. Code, models and pre-computed image features are available from the project website\footnote{http://www.panderson.me/up-down-attention}.
 
  
\section{Related Work}

A large number of attention-based deep neural networks have been proposed for image captioning and VQA. Typically, these models can be characterized as top-down approaches, with context provided by a representation of a partially-completed caption in the case of image captioning~\cite{scst2016,sentinel,reviewnet,Xu2015}, or a representation of the question in the case of VQA~\cite{fukui2016multimodal,coatt,askattend,stacked,visual7w}. In each case attention is applied to the output of one or more layers of a CNN, by predicting a weighting for each spatial location in the CNN output. However, determining the optimal number of image regions invariably requires an unwinnable trade-off between coarse and fine levels of detail. Furthermore, the arbitrary positioning of the regions with respect to image content may make it more difficult to detect objects that are poorly aligned to regions and to bind visual concepts associated with the same object. 

Comparatively few previous works have considered applying attention to salient image regions. We are aware of two papers. Jin et al.~\cite{factorization} use selective search~\cite{selective} to identify salient image regions, which are filtered with a classifier then resized and CNN-encoded as input to an image captioning model with attention. The Areas of Attention captioning model~\cite{areas} uses either edge boxes~\cite{edge} or spatial transformer networks~\cite{stn} to generate image features, which are processed using an attention model based on three bi-linear pairwise interactions~\cite{areas}. In this work, rather than using hand-crafted or differentiable region proposals~\cite{selective,edge,stn}, we leverage Faster R-CNN~\cite{faster_rcnn}, establishing a closer link between vision and language tasks and recent progress in object detection. With this approach we are able to pre-train our region proposals on object detection datasets. Conceptually, the advantages should be similar to pre-training visual representations on ImageNet~\cite{ILSVRC15} and leveraging significantly larger cross-domain knowledge. We additionally apply our method to VQA, establishing the broad applicability of our approach.


\section{Approach}
\label{sec:approach}

Given an image $I$, both our image captioning model and our VQA model take as input a possibly variably-sized set of $k$ image features, $V = \{\bv_1,...,\bv_k\}, \bv_i \in \mathbb{R}^D$, such that each image feature encodes a salient region of the image. The spatial image features $V$ can be variously defined as the output of our bottom-up attention model, or, following standard practice, as the spatial output layer of a CNN. We describe our approach to implementing a bottom-up attention model in \secref{sec:rcnn}. In \secref{sec:cap} we outline the architecture of our image captioning model and in \secref{sec:vqa} we outline our VQA model. We note that for the top-down attention component, both models use simple one-pass attention mechanisms, as opposed to the more complex schemes of recent models such as stacked, multi-headed, or bidirectional attention~\cite{stacked,jabri2016revisiting,kazemi2017baseline,coatt} that could also be applied.

\begin{figure}[t]
	\begin{center}
		\includegraphics[width=1.0\linewidth]{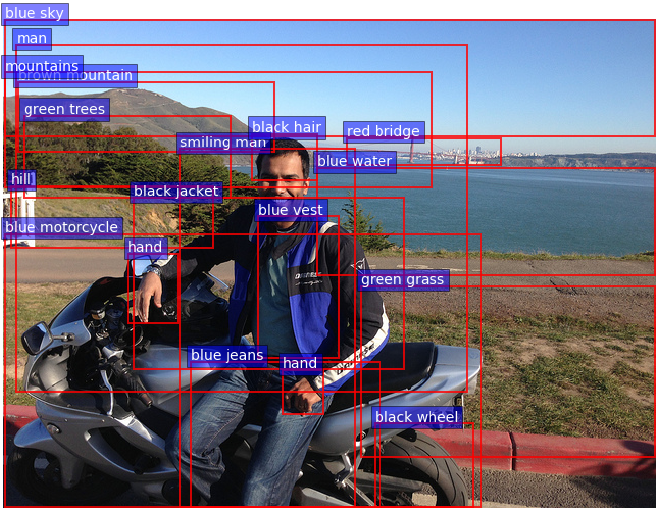}\\
		\includegraphics[width=1.0\linewidth]{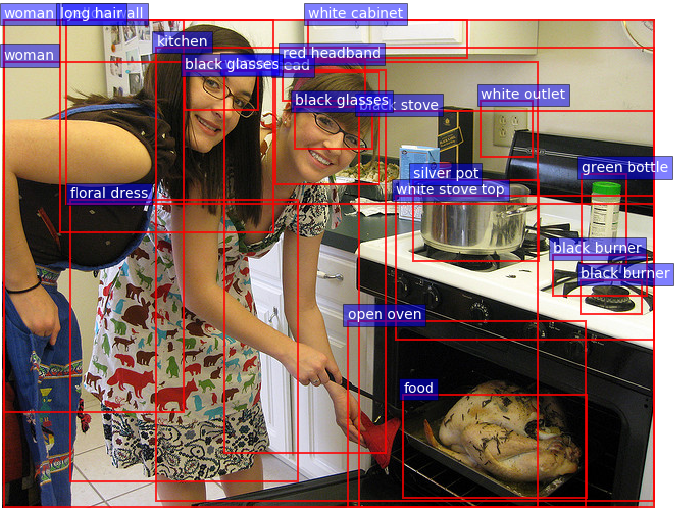}\\
	\end{center}
	\caption{Example output from our Faster R-CNN bottom-up attention model. Each bounding box is labeled with an attribute class followed by an object class. Note however, that in captioning and VQA we utilize only the feature vectors -- not the predicted labels.}
	\label{fig:rcnn-demo}
\end{figure}

\subsection{Bottom-Up Attention Model}
\label{sec:rcnn}

The definition of spatial image features $V$ is generic. However, in this work we define spatial regions in terms of bounding boxes and implement bottom-up attention using Faster R-CNN~\cite{faster_rcnn}. Faster R-CNN is an object detection model designed to identify instances of objects belonging to certain classes and localize them with bounding boxes. Other region proposal networks could also be trained as an attentive mechanism~\cite{yolo2016,liu2015ssd}.

Faster R-CNN detects objects in two stages. The first stage, described as a Region Proposal Network (RPN), predicts object proposals. A small network is slid over features at an intermediate level of a CNN. At each spatial location the network predicts a class-agnostic objectness score and a bounding box refinement for anchor boxes of multiple scales and aspect ratios. Using greedy non-maximum suppression with an intersection-over-union (IoU) threshold, the top box proposals are selected as input to the second stage. In the second stage, region of interest (RoI) pooling is used to extract a small feature map (e.g. $14\times14$) for each box proposal. These feature maps are then batched together as input to the final layers of the CNN. The final output of the model consists of a softmax distribution over class labels and class-specific bounding box refinements for each box proposal. 

In this work, we use Faster R-CNN in conjunction with the ResNet-101~\cite{he2015deep} CNN. To generate an output set of image features $V$ for use in image captioning or VQA, we take the final output of the model and perform non-maximum suppression for each object class using an IoU threshold. We then select all regions where any class detection probability exceeds a confidence threshold. 
For each selected region $i$, $\bv_i$ is defined as the mean-pooled convolutional feature from this region, such that the dimension $D$ of the image feature vectors is 2048. Used in this fashion, Faster R-CNN effectively functions as a `hard' attention mechanism, as only a relatively small number of image bounding box features are selected from a large number of possible configurations. 

To pretrain the bottom-up attention model, we first initialize Faster R-CNN with ResNet-101 pretrained for classification on ImageNet~\cite{ILSVRC15}. We then train on Visual Genome~\cite{krishnavisualgenome} data. To aid the learning of good feature representations, we add an additional training output for predicting attribute classes (in addition to object classes). To predict attributes for region $i$, we concatenate the mean pooled convolutional feature $\bv_i$ with a learned embedding of the ground-truth object class, and feed this into an additional output layer defining a softmax distribution over each attribute class plus a `no attributes' class.

The original Faster R-CNN multi-task loss function contains four components, defined over the classification and bounding box regression outputs for both the RPN and the final object class proposals respectively. We retain these components and add an additional multi-class loss component to train the attribute predictor. In \figref{fig:rcnn-demo} we provide some examples of model output. 

\subsection{Captioning Model}
\label{sec:cap}

Given a set of image features $V$, our proposed captioning model uses a `soft' top-down attention mechanism to weight each feature during caption generation, using the existing partial output sequence as context. This approach is broadly similar to several previous works~\cite{scst2016,sentinel,Xu2015}. However, the particular design choices outlined below make for a relatively simple yet high-performing baseline model. Even without bottom-up attention, our captioning model achieves performance comparable to state-of-the-art on most evaluation metrics (refer \tabref{tab:karpathy}).

\begin{figure}[t]
	\begin{center}
		\includegraphics[width=0.95\linewidth]{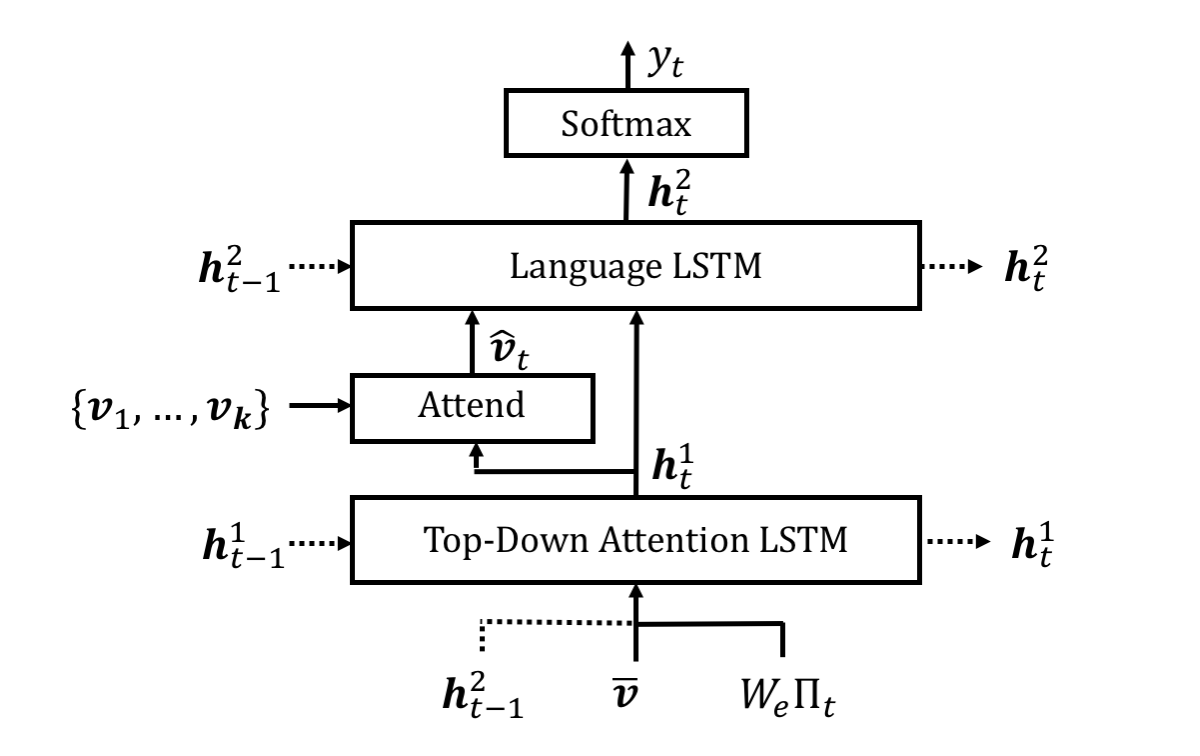}
	\end{center}
	\caption{Overview of the proposed captioning model. Two LSTM layers are used to selectively attend to spatial image features $\{\bv_1,...,\bv_k\}$. These features can be defined as the spatial output of a CNN, or following our approach, generated using bottom-up attention.}
	\label{fig:captioner}
\end{figure}

At a high level, the captioning model is composed of two LSTM~\cite{Hochreiter1997} layers using a standard implementation~\cite{Donahue2015}. In the sections that follow we will refer to the operation of the LSTM over a single time step using the following notation:
\begin{align}
\bh_t &= \textrm{LSTM}(\bx_t, \bh_{t-1})
\end{align}
\noindent
where $\bx_t$ is the LSTM input vector and $\bh_t$ is the LSTM output vector. Here we have neglected the propagation of memory cells for notational convenience. We now describe the formulation of the LSTM input vector $\bx_t$ and the output vector $\bh_t$ for each layer of the model. The overall captioning model is illustrated in \figref{fig:captioner}. 

\subsubsection{Top-Down Attention LSTM}

Within the captioning model, we characterize the first LSTM layer as a top-down visual attention model, and the second LSTM layer as a language model, indicating each layer with superscripts in the equations that follow. Note that the bottom-up attention model is described in \secref{sec:rcnn}, and in this section its outputs are simply considered as features $V$. The input vector to the attention LSTM at each time step consists of the previous output of the language LSTM, concatenated with the mean-pooled image feature $\bar{\bv} = \frac{1}{k}\sum_i \bv_i$ and an encoding of the previously generated word, given by:
\begin{align}
\bx_t^1 &= [\bh_{t-1}^2, \bar{\bv}, W_{e}\Pi_t]
\end{align}
\noindent where $W_{e} \in \mathbb{R}^{E \times \vert\Sigma\vert}$ is a word embedding matrix for a vocabulary $\Sigma$, and $\Pi_t $ is one-hot encoding of the input word at timestep $t$. These inputs provide the attention LSTM with maximum context regarding the state of the language LSTM, the overall content of the image, and the partial caption output generated so far, respectively. The word embedding is learned from random initialization without pretraining.

Given the output $\bh_t^1$ of the attention LSTM, at each time step $t$ we generate a normalized attention weight $\alpha_{i,t}$ for each of the $k$ image features $\bv_i$ as follows:
\begin{align}
\label{eqn:att}
a_{i,t} &= \bw_a^T \tanh\,(W_{va}\bv_i + W_{ha}\bh_t^1) \\
\label{eqn:norm}
\boldsymbol{\alpha}_{t} &= \textrm{softmax}\,(\ba_{t})
\end{align}
\noindent
where $W_{va} \in \mathbb{R}^{H\times V}$, $W_{ha} \in \mathbb{R}^{H\times M}$ and $\bw_a \in \mathbb{R}^{H}$ are learned parameters. The attended image feature used as input to the language LSTM is calculated as a convex combination of all input features:
\begin{align}
\label{eqn:attended}
\hat{\bv}_t &= \sum_{i=1}^{K}{\alpha_{i,t}\bv_i}
\end{align}

\subsubsection{Language LSTM}

The input to the language model LSTM consists of the attended image feature, concatenated with the output of the attention LSTM, given by:
\begin{align}
\bx_t^2 &= [\hat{\bv}_t, \bh_{t}^1]
\end{align}
Using the notation $y_{1:T}$ to refer to a sequence of words $(y_1, ..., y_T)$, at each time step $t$ the conditional distribution over possible output words is given by:
\begin{align}
p(y_t \mid y_{1:t-1}) &= \textrm{softmax}\,(W_{p}\bh_{t}^2 + \bb_p)
\end{align}
\noindent
where $W_{p} \in \mathbb{R}^{\vert\Sigma\vert \times M}$ and $\bb_p \in \mathbb{R}^{\vert\Sigma\vert}$ are learned weights and biases. The distribution over complete output sequences is calculated as the product of conditional distributions:
\begin{align}
p(y_{1:T}) &= \prod_{t=1}^{T} p(y_t \mid y_{1:t-1})
\end{align}

\subsubsection{Objective}

\begin{figure*}
	\begin{center}
		\includegraphics[width=1\linewidth]{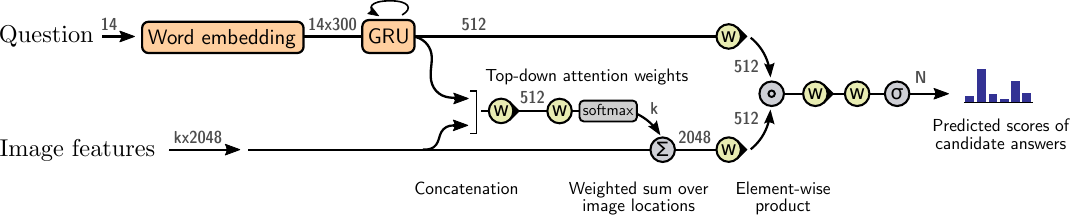}
	\end{center}
\caption{Overview of the proposed VQA model. A deep neural network implements a joint embedding of the question and image features $\{\bv_1,...,\bv_k\}$ . These features can be defined as the spatial output of a CNN, or following our approach, generated using bottom-up attention. Output is generated by a multi-label classifier operating over a fixed set of candidate answers. Gray numbers indicate the dimensions of the vector representations between layers. Yellow elements use learned parameters.}
	\label{fig:vqa}
\end{figure*}

Given a target ground truth sequence $y_{1:T}^*$ and a captioning model with parameters $\theta$, we minimize the following cross entropy loss:
\begin{align}
L_{XE}(\theta) &= -\sum_{t=1}^{T}\log(p_{\theta}(y_t^* \mid y_{1:t-1}^*))
\end{align}

For fair comparison with recent work~\cite{scst2016} we also report results optimized for CIDEr~\cite{Vedantam2015}. Initializing from the cross-entropy trained model, we seek to minimize the negative expected score:
\begin{align}
\label{eqn:ciderloss}
L_{R}(\theta) &= -\expectwrt[y_{1:T} \sim p_\theta]{r(y_{1:T})}
\end{align}
\noindent
where $r$ is the score function (e.g., CIDEr). Following the approach described as Self-Critical Sequence Training~\cite{scst2016} (SCST), the gradient of this loss can be approximated:
\begin{align}
\nabla_{\theta}L_{R}(\theta) \approx -(r(y_{1:T}^s)-r(\hat{y}_{1:T}))\nabla_{\theta}\log p_\theta(y_{1:T}^s)
\end{align}
\noindent where $y_{1:T}^s$ is a sampled caption and $r(\hat{y}_{1:T})$ defines the baseline score obtained by greedily decoding the current model. SCST (like other REINFORCE~\cite{Williams:1992:SSG:139611.139614} algorithms) explores the space of captions by sampling from the policy during training. This gradient tends to increase the probability of sampled captions that score higher than the score from the current model. 

In our experiments, we follow SCST but we speed up the training process by restricting the sampling distribution. Using beam search decoding, we sample only from those captions in the decoded beam. Empirically, we have observed when decoding using beam search that the resulting beam typically contains at least one very high scoring caption -- although frequently this caption does not have the highest log-probability of the set. In contrast, we observe that very few unrestricted caption samples score higher than the greedily-decoded caption. Using this approach, we complete CIDEr optimization in a single epoch.

\subsection{VQA Model}
\label{sec:vqa}

Given a set of spatial image features $V$, our proposed VQA model also uses a `soft' top-down attention mechanism to weight each feature, using the question representation as context. As illustrated in \figref{fig:vqa}, the proposed model implements the well-known joint multimodal embedding of the question and the image, followed by a prediction of regression of scores over a set of candidate answers. This approach has been the basis of numerous previous models~\cite{jabri2016revisiting,kazemi2017baseline,teney2016zsvqa}. However, as with our captioning model, implementation decisions are important to ensure that this relatively simple model delivers high performance.

The learned non-linear transformations within the network are implemented with gated hyperbolic tangent activations~\cite{dauphin2016languagecnn}. These are a special case of highway networks \cite{srivastava2015highway} that have shown a strong empirical advantage over traditional ReLU or tanh layers. Each of our `gated tanh' layers implements a function $f_a:\bx \in \mathbb{R}^{m} \to \by \in \mathbb{R}^{n}$ with parameters $a = \{W, W', b, b'\}$ defined as follows:
\begin{align}
  \label{eq:nonLinearBlock}
  & \tilde{\by} = \tanh\,(W \bx + \bb )\\
  & \boldsymbol{g} = \sigma( W' \bx + \bb' )\\
  & \by = \tilde{\by} ~\circ~ \boldsymbol{g}
\end{align}
\noindent
where $\sigma$ is the sigmoid activation function, $W, W' \in \mathbb{R}^{n \times m}$ are learned weights, $\bb, \bb' \in \mathbb{R}^{n}$ are learned biases, and $\circ$ is the Hadamard (element-wise) product. The vector $\boldsymbol{g}$ acts multiplicatively as a gate on the intermediate activation $\tilde{\by}$.

Our proposed approach first encodes each question as the hidden state $\bq$ of a gated recurrent unit~\cite{cho2014learning} (GRU), with each input word represented using a learned word embedding. Similar to \eqnref{eqn:att}, given the output $\bq$ of the GRU, we generate an unnormalized attention weight $a_{i}$ for each of the $k$ image features $\bv_i$ as follows:
\begin{align}
\label{eqn:att-vqa}
a_{i} &= \bw_a^T  f_a( [\bv_i,\bq])
\end{align}
\noindent
where $\bw_a^T$ is a learned parameter vector. \eqnref{eqn:norm} and \eqnref{eqn:attended} (neglecting subscripts $t$) are used to calculate the normalized attention weight and the attended image feature $\hat{\bv}$. The distribution over possible output responses $y$ is given by:
\begin{align}
\bh &= f_q(\bq) ~\circ~ f_v(\hat{\bv})\\
p(y) &= \sigma(W_{o} \, f_o(\bh) )
\end{align}
Where $\bh$ is a joint representation of the question and the image, and $W_{o} \in \mathbb{R}^{\vert\Sigma\vert \times M}$ are learned weights. 

Due to space constraints, some important aspects of our VQA approach are not detailed here. For full specifics of the VQA model including a detailed exploration of architectures and hyperparameters, refer to Teney et al.~\cite{teney2017tips}.

\section{Evaluation}

\begin{table*}
	\begin{center}\small
		\setlength{\tabcolsep}{.16em}
		\begin{tabular}{lccccccccccccc}
			\midrule
			& \multicolumn{6}{c}{Cross-Entropy Loss}     &                                                    & \multicolumn{6}{c}{CIDEr Optimization}                                                       \\
			\cmidrule{2-7}
			\cmidrule{9-14}
			& BLEU-1        & BLEU-4        & METEOR        & ROUGE-L       & CIDEr          & SPICE   &  \:\:    & BLEU-1        & BLEU-4        & METEOR        & ROUGE-L       & CIDEr          & SPICE         \\
			\midrule
			SCST:Att2in~\cite{scst2016}                &  -             & 31.3          & 26.0            & 54.3          & 101.3          &  -             &&   -          & 33.3          & 26.3          & 55.3          & 111.4          &  -             \\
			SCST:Att2all~\cite{scst2016}                &  -             & 30.0          & 25.9            & 53.4          & 99.4          &  -             &&   -          & 34.2          & 26.7          & 55.7          & 114.0          &  -             \\
			\midrule
			Ours: ResNet        & 74.5          & 33.4          & 26.1          & 54.4          & 105.4          & 19.2          && 76.6          & 34.0          & 26.5          & 54.9          & 111.1          & 20.2          \\
			Ours: Up-Down           & \textbf{77.2} & \textbf{36.2} & \textbf{27.0} & \textbf{56.4} & \textbf{113.5} & \textbf{20.3} && \textbf{79.8} & \textbf{36.3} & \textbf{27.7} & \textbf{56.9} & \textbf{120.1} & \textbf{21.4} \\
			Relative Improvement      & 4\%       & 8\%          & 3\%          & 4\%          & 8\%          & 6\%        & & 4\%       & 7\%          & 5\%          & 4\%          & 8\%          & 6\%   \\
			\midrule
		\end{tabular}
	\end{center}
	\caption{Single-model image captioning performance on the MSCOCO Karpathy test split. Our baseline ResNet model obtains similar results to SCST~\cite{scst2016}, the existing state-of-the-art on this test set. Illustrating the contribution of bottom-up attention, our Up-Down model achieves significant (3--8\%) relative gains across all metrics regardless of whether cross-entropy loss or CIDEr optimization is used. }
	\label{tab:karpathy}
\end{table*}

\begin{table*}
	\begin{center}\small
		\setlength{\tabcolsep}{.25em}
		\begin{tabular}{lccccccccccccccc}
			\midrule
			& \multicolumn{7}{c}{Cross-Entropy Loss}     &                                                    & \multicolumn{7}{c}{CIDEr Optimization}                                                       \\
			\cmidrule{2-8}
			\cmidrule{10-16}
			& SPICE  & Objects        & Attributes        & Relations       & Color       & Count          & Size   &  \:\:    & SPICE & Objects        & Attributes        & Relations        & Color       & Count          & Size         \\
			\midrule
			Ours: ResNet        & 19.2       & 35.4          & 8.6          & 5.3          & 12.2          & 4.1       & 3.9  & & 20.2          & 37.0          & 9.2          & 6.1          & 10.6          & 12.0   &   \textbf{4.3}    \\
			Ours: Up-Down    &   \textbf{20.3}    & \textbf{37.1} & \textbf{9.2} & \textbf{5.8} & \textbf{12.7} & \textbf{6.5} & \textbf{4.5} && \textbf{21.4} & \textbf{39.1} & \textbf{10.0} & \textbf{6.5} & \textbf{11.4} & \textbf{18.4}  & 3.2 \\
			\midrule
		\end{tabular}
	\end{center}
	\caption{Breakdown of SPICE F-scores over various subcategories on the MSCOCO Karpathy test split. Our Up-Down model outperforms the ResNet baseline at identifying objects, as well as detecting object attributes and the relations between objects. }
	\label{tab:spice}
\end{table*}

\subsection{Datasets}

\subsubsection{Visual Genome Dataset}

We use the Visual Genome~\cite{krishnavisualgenome} dataset to pretrain our bottom-up attention model, and for data augmentation when training our VQA model. The dataset contains 108K images densely annotated with scene graphs containing objects, attributes and relationships, as well as 1.7M visual question answers.

For pretraining the bottom-up attention model, we use only the object and attribute data. We reserve 5K images for validation, and 5K images for future testing, treating the remaining 98K images as training data. As approximately 51K Visual Genome images are also found in the MSCOCO captions dataset~\cite{Lin2014}, we are careful to avoid contamination of our MSCOCO validation and test sets. We ensure that any images found in both datasets are contained in the same split in both datasets. 

As the object and attribute annotations consist of freely annotated strings, rather than classes, we perform extensive cleaning and filtering of the training data. Starting from 2,000 object classes and 500 attribute classes, we manually remove abstract classes that exhibit poor detection performance in initial experiments. Our final training set contains 1,600 object classes and 400 attribute classes. Note that we do not merge or remove overlapping classes (e.g. `person', `man', `guy'), classes with both singular and plural versions (e.g. `tree', `trees') and classes that are difficult to precisely localize (e.g. `sky', `grass', `buildings').

When training the VQA model, we augment the VQA v2.0 training data with Visual Genome question and answer pairs provided the correct answer is present in model's answer vocabulary. This represents about 30\% of the available data, or 485K questions.

\subsubsection{Microsoft COCO Dataset}

To evaluate our proposed captioning model, we use the MSCOCO 2014 captions dataset~\cite{Lin2014}. For validation of model hyperparameters and offline testing, we use the `Karpathy' splits~\cite{neuraltalk} that have been used extensively for reporting results in prior work. This split contains 113,287 training images with five captions each, and 5K images respectively for validation and testing. Our MSCOCO test server submission is trained on the entire MSCOCO 2014 training and validation set (123K images). 

We follow standard practice and perform only minimal text pre-processing, converting all sentences to lower case, tokenizing on white space, and filtering words that do not occur at least five times, resulting in a model vocabulary of 10,010 words. To evaluate caption quality, we use the standard automatic evaluation metrics, namely SPICE~\cite{spice2016}, CIDEr~\cite{Vedantam2015}, METEOR~\cite{meteor-wmt:2014}, ROUGE-L~\cite{Lin2004} and BLEU~\cite{Papineni2002}.

\subsubsection{VQA v2.0 Dataset}

To evaluate our proposed VQA model, we use the recently introduced VQA v2.0 dataset~\cite{balanced_vqa_v2}, which attempts to minimize the effectiveness of learning dataset priors by balancing the answers to each question. The dataset, which was used as the basis of the 2017 VQA Challenge\footnote{http://www.visualqa.org/challenge.html}, contains 1.1M questions with 11.1M answers relating to MSCOCO images. 

We perform standard question text preprocessing and tokenization. Questions are trimmed to a maximum of 14 words for computational efficiency. The set of candidate answers is restricted to correct answers in the training set that appear more than 8 times, resulting in an output vocabulary size of 3,129. Our VQA test server submissions are trained on the training and validation sets plus additional questions and answers from Visual Genome. To evaluate answer quality, we report accuracies using the standard VQA metric~\cite{VQA}, which takes into account the occasional disagreement between annotators for the ground truth answers.

\begin{table*}
	\begin{center}\small
		\setlength{\tabcolsep}{.35em}
		\begin{tabular}{lccccccccccccccccccccccc}
			\midrule
			& \multicolumn{2}{c}{BLEU-1} &  & \multicolumn{2}{c}{BLEU-2} &  & \multicolumn{2}{c}{BLEU-3} &  & \multicolumn{2}{c}{BLEU-4} &  & \multicolumn{2}{c}{METEOR} &  & \multicolumn{2}{c}{ROUGE-L} &  & \multicolumn{2}{c}{CIDEr}  &  & \multicolumn{2}{c}{SPICE} \\
			\cmidrule{2-3}\cmidrule{5-6}\cmidrule{8-9}\cmidrule{11-12}\cmidrule{14-15}\cmidrule{17-18}\cmidrule{20-21}\cmidrule{23-24}
			& c5           & c40         &  & c5           & c40         &  & c5           & c40         &  & c5           & c40         &  & c5           & c40         &  & c5           & c40         &  & c5           & c40         &  & c5           & c40         \\
			\midrule
			Review Net~\cite{reviewnet}&72.0&90.0& &55.0&81.2& &41.4&70.5& &31.3&59.7& &25.6&34.7& &53.3&68.6& &96.5&96.9& &18.5&64.9\\
			Adaptive~\cite{sentinel}&74.8&92.0& &58.4&84.5& &44.4&74.4& &33.6&63.7& &26.4&35.9& &55.0&70.5& &104.2&105.9& &19.7&67.3\\
			PG-BCMR~\cite{LiuZYG017}&75.4&-& &59.1&-& &44.5&-& &33.2&-& &25.7&-& &55&-& &101.3&-& &-&-\\
			SCST:Att2all~\cite{scst2016}&78.1&93.7& &61.9&86.0& &47.0&75.9& &35.2&64.5& &27.0&35.5& &56.3&70.7& &114.7&116.7& &20.7&68.9\\
			LSTM-A$_{\text{3}}$~\cite{yao-msm}&78.7&93.7& &62.7&86.7& &47.6&76.5& &35.6&65.2& &27&35.4& &56.4&70.5& &116&118& &-&-\\
			Ours: Up-Down & \textbf{80.2} & \textbf{95.2} &  & \textbf{64.1} & \textbf{88.8} &  & \textbf{49.1} & \textbf{79.4} &  & \textbf{36.9} & \textbf{68.5} &  & \textbf{27.6} & \textbf{36.7} &  & \textbf{57.1} & \textbf{72.4} &  & \textbf{117.9} & \textbf{120.5} &  & \textbf{21.5} & \textbf{71.5} \\
			\midrule
		\end{tabular}
	\end{center}
	\caption{Highest ranking published image captioning results on the online MSCOCO test server. Our submission, an ensemble of 4 models optimized for CIDEr with different initializations, outperforms previously published work on all reported metrics. At the time of submission (18 July 2017), we also outperformed all unpublished test server submissions.}
	\label{tab:server}
\end{table*}

\begin{figure*}
	\begin{center}
		\includegraphics[width=1\linewidth]{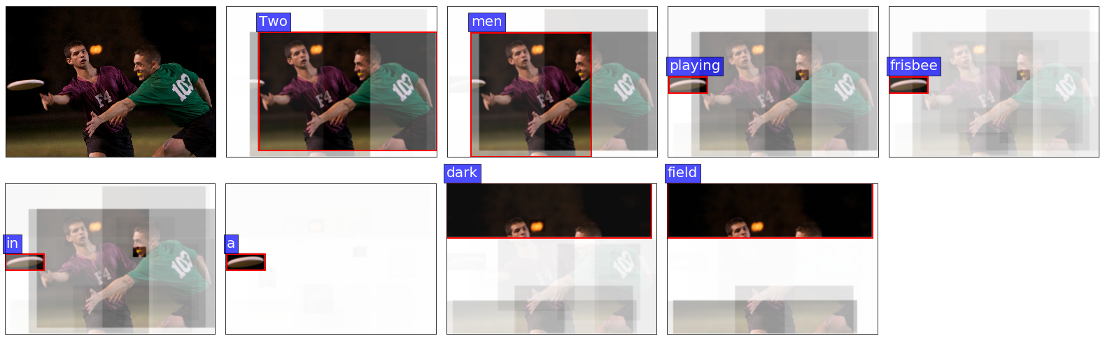}\\
		Two men playing frisbee in a dark field.
	\end{center}
	\caption{Example of a generated caption showing attended image regions. For each generated word, we visualize the attention weights on individual pixels, outlining the region with the maximum attention weight in red. Avoiding the conventional trade-off between coarse and fine levels of detail, our model focuses on both closely-cropped details, such as the frisbee and the green player's mouthguard when generating the word `playing', as well as large regions, such as the night sky when generating the word `dark'.  }
	\label{fig:caption_example}
\end{figure*}

\subsection{ResNet Baseline}

To quantify the impact of bottom-up attention, in both our captioning and VQA experiments we evaluate our full model (\textit{Up-Down}) against prior work as well as an ablated baseline. In each case, the baseline (\textit{ResNet}), uses a ResNet~\cite{he2015deep} CNN pretrained on ImageNet~\cite{ILSVRC15} to encode each image in place of the bottom-up attention mechanism. 

In image captioning experiments, similarly to previous work~\cite{scst2016} we encode the full-sized input image with the final convolutional layer of Resnet-101, and use bilinear interpolation to resize the output to a fixed size spatial representation of 10$\times$10. This is equivalent to the maximum number of spatial regions used in our full model. In VQA experiments, we encode the resized input image with ResNet-200~\cite{he2016identityresnet}. In separate experiments we use evaluate the effect of varying the size of the spatial output from its original size of 14$\times$14, to 7$\times$7 (using bilinear interpolation) and 1$\times$1 (i.e., mean pooling without attention).


\subsection{Image Captioning Results}

In \tabref{tab:karpathy} we report the performance of our full model and the ResNet baseline in comparison to the existing state-of-the-art Self-critical Sequence Training~\cite{scst2016} (SCST) approach on the test portion of the Karpathy splits. For fair comparison, results are reported for models trained with both standard cross-entropy loss, and models optimized for CIDEr score. Note that the SCST approach uses ResNet-101 encoding of full images, similar to our ResNet baseline. All results are reported for a single model with no fine-tuning of the input ResNet / R-CNN model. However, the SCST results are selected from the best of four random initializations, while our results are outcomes from a single initialization.

Relative to the SCST models, our ResNet baseline obtains slightly better performance under cross-entropy loss, and slightly worse performance when optimized for CIDEr score. After incorporating bottom-up attention, our full Up-Down model shows significant improvements across all metrics regardless of whether cross-entropy loss or CIDEr optimization is used. Using just a single model, we obtain the best reported results for the Karpathy test split. As illustrated in \tabref{tab:spice}, the contribution from bottom-up attention is broadly based, illustrated by improved performance in terms of identifying objects, object attributes and also the relationships between objects.

\tabref{tab:server} reports the performance of 4 ensembled models trained with CIDEr optimization on the official MSCOCO evaluation server, along with the highest ranking previously published results. At the time of submission (18 July 2017), we outperform all other test server submissions on all reported evaluation metrics.

\subsection{VQA Results}

\begin{table}[t]
\small
\centering
\setlength{\tabcolsep}{.5em}
\begin{tabular}{llcccc}
\midrule
                                    &    &   Yes/No  &  Number   &  Other & Overall  \\
\midrule
Ours: ResNet (1$\times$1) &    			 & 76.0	& 36.5	& 46.8 & 56.3	\\
Ours: ResNet (14$\times$14) &   				 & 76.6	& 36.2	& 49.5	& 57.9 \\
Ours: ResNet (7$\times$7) &			 & 77.6	& 37.7	& 51.5 & 59.4	\\
Ours: Up-Down  &  & \textbf{80.3} & \textbf{42.8} & \textbf{55.8} & \textbf{63.2}\\
Relative Improvement &			 & 3\%	& 14\%	& 8\% & 6\%	\\
\midrule
\end{tabular}
\caption{Single-model performance on the VQA v2.0 validation set. The use of bottom-up attention in the Up-Down model provides a significant improvement over the best ResNet baseline across all question types, even though the ResNet baselines use almost twice as many convolutional layers.}
\label{tab:vqa_val}
\end{table}

\begin{table}[t]
\small
\centering
\setlength{\tabcolsep}{.5em}
\begin{tabular}{llcccc}
\midrule
                                    &    &   Yes/No  &  Number   &  Other & Overall  \\
\midrule
Prior~\cite{balanced_vqa_v2} &						 & 61.20	& 0.36	& 1.17	& 25.98\\
Language-only~\cite{balanced_vqa_v2} &									 & 67.01	& 31.55	& 27.37	& 44.26\\
d-LSTM+n-I~\cite{lu2015deeperLstm,balanced_vqa_v2} &		 & 73.46	& 35.18	& 41.83 & 54.22	\\
MCB~\cite{fukui2016multimodal,balanced_vqa_v2} &					 & 78.82	& 38.28	& 53.36	& 62.27\\
UPMC-LIP6 &    			 & 82.07	& 41.06	& 57.12 & 65.71	\\
Athena &   				 & 82.50	& 44.19	& 59.97	& 67.59 \\
HDU-USYD-UNCC &			 & 84.50	& 45.39	& 59.01 & 68.09	\\
Ours: Up-Down  &  & \textbf{86.60} & \textbf{48.64} & \textbf{61.15} & \textbf{70.34}\\
\midrule
\end{tabular}
\caption{VQA v2.0 test-standard server accuracy as at 8 August 2017, ranking our submission against published and unpublished work for each question type. Our approach, an ensemble of 30 models, outperforms all other leaderboard entries.}
\label{tab:vqa_test}
\end{table}

In \tabref{tab:vqa_val} we report the single model performance of our full Up-Down VQA model relative to several ResNet baselines on the VQA v2.0 validation set. The addition of bottom-up attention provides a significant improvement over the best ResNet baseline across all question types, even though the ResNet baseline uses approximately twice as many convolutional layers.  \tabref{tab:vqa_test} reports the performance of 30 ensembled models on the official VQA 2.0 test-standard evaluation server, along with the previously published baseline results and the highest ranking other entries. At the time of submission (8 August 2017), we outperform all other test server submissions. Our submission also achieved first place in the 2017 VQA Challenge.

\subsection{Qualitative Analysis}

To help qualitatively evaluate our attention methodology, in \figref{fig:caption_example} we visualize the attended image regions for different words generated by our Up-Down captioning model. As indicated by this example, our approach is equally capable of focusing on fine details or large image regions. This capability arises because the attention candidates in our model consist of many overlapping regions with varying scales and aspect ratios -- each aligned to an object, several related objects, or an otherwise salient image patch. 

\begin{figure}[t]
	\begin{center}
		\includegraphics[width=1\linewidth]{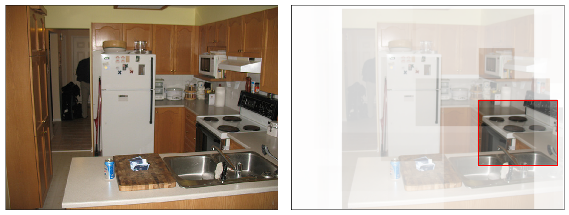}
		Question: What room are they in?
		Answer: kitchen
	\end{center}
\caption{VQA example illustrating attention output. Given the question `What room are they in?', the model focuses on the stovetop, generating the answer `kitchen'. }
	\label{fig:vqa_example}
\end{figure}

Unlike conventional approaches, when a candidate attention region corresponds to an object, or several related objects, all the visual concepts associated with those objects appear to be spatially co-located -- and are processed together. In other words, our approach is able to consider all of the information pertaining to an object at once. This is also a natural way for attention to be implemented. In the human visual system, the problem of integrating the separate features of objects in the correct combinations is known as the feature binding problem, and experiments suggest that attention plays a central role in the solution~\cite{TreismanGelade80,treisman1982perceptual}. We include an example of VQA attention in \figref{fig:vqa_example}.

\section{Conclusion}

We present a novel combined bottom-up and top-down visual attention mechanism. Our approach enables attention to be calculated more naturally at the level of objects and other salient regions. Applying this approach to image captioning and visual question answering, we achieve state-of-the-art results in both tasks, while improving the interpretability of the resulting attention weights. 

At a high level, our work more closely unifies tasks involving visual and linguistic understanding with recent progress in object detection. While this suggests several directions for future research, the immediate benefits of our approach may be captured by simply replacing pretrained CNN features with pretrained bottom-up attention features.  

\subsection*{Acknowledgements}
\small
\noindent
This research is partially supported by an Australian Government Research Training Program (RTP) Scholarship, by the Australian Research Council Centre of Excellence for Robotic Vision (project number CE140100016), by a Google award through the Natural Language Understanding Focused Program, and under the Australian Research Council’s Discovery Projects funding scheme (project number DP160102156).

{\small
\bibliographystyle{ieee}
\bibliography{paper}
}

\newpage

\section*{SUPPLEMENTARY MATERIALS}

\section{Implementation Details}

\subsection{Bottom-Up Attention Model}

Our bottom-up attention Faster R-CNN implementation uses an IoU threshold of 0.7 for region proposal suppression, and 0.3 for object class suppression. To select salient image regions, a class detection confidence threshold of 0.2 is used, allowing the number of regions per image $k$ to vary with the complexity of the image, up to a maximum of 100. However, in initial experiments we find that simply selecting the top 36 features in each image works almost as well in both downstream tasks. Since Visual Genome~\cite{krishnavisualgenome} contains a relatively large number of annotations per image, the model is relatively intensive to train. Using 8 Nvidia M40 GPUs, we take around 5 days to complete 380K training iterations, although we suspect that faster training regimes could also be effective.

\subsection{Captioning Model}

In the captioning model, we set the number of hidden units $M$ in each LSTM to 1,000, the number of hidden units $H$ in the attention layer to 512, and the size of the input word embedding $E$ to 1,000. In training, we use  a simple learning rate schedule, beginning with a learning rate of 0.01 which is reduced to zero on a straight-line basis over 60K iterations using a batch size of 100 and a momentum parameter of 0.9. Training using two Nvidia Titan X GPUs takes around 9 hours (including less than one hour for CIDEr optimization). During optimization and decoding we use a beam size of 5. When decoding we also enforce the constraint that a single word cannot be predicted twice in a row. Note that in both our captioning and VQA models, image features are fixed and not finetuned. 

\subsection{VQA Model}

In the VQA model, we use 300 dimension word embeddings, initialized with pretrained GloVe vectors~\cite{pennington2014glove}, and we use hidden states of dimension 512. We train the VQA model using AdaDelta~\cite{zeiler2012adadelta} and regularize with early stopping. The training of the model takes in the order of 12--18 hours on a single Nvidia K40 GPU. Refer to Teney et al.~\cite{teney2017tips} for further details of the VQA model implementation.

\section{Additional Examples}

\begin{figure*}
	\begin{center}
		Resnet -- A man sitting on a \textit{toilet} in a bathroom.
		\includegraphics[width=1\linewidth]{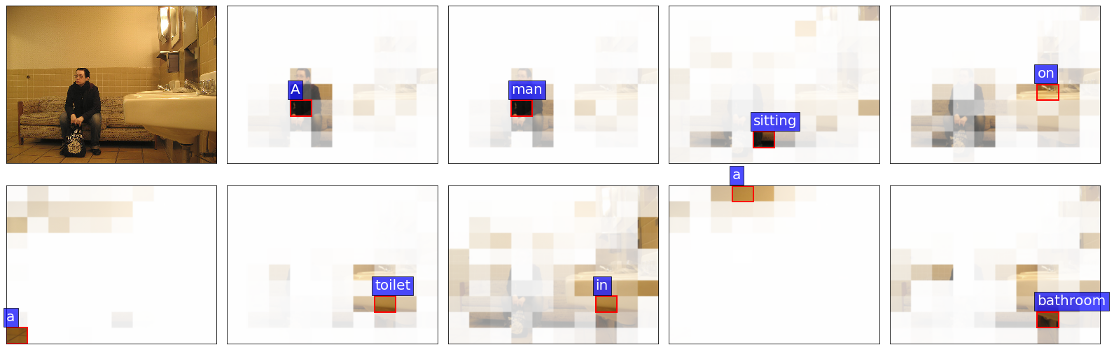}\\
		Up-Down -- A man sitting on a \textit{couch} in a bathroom.
		\includegraphics[width=1\linewidth]{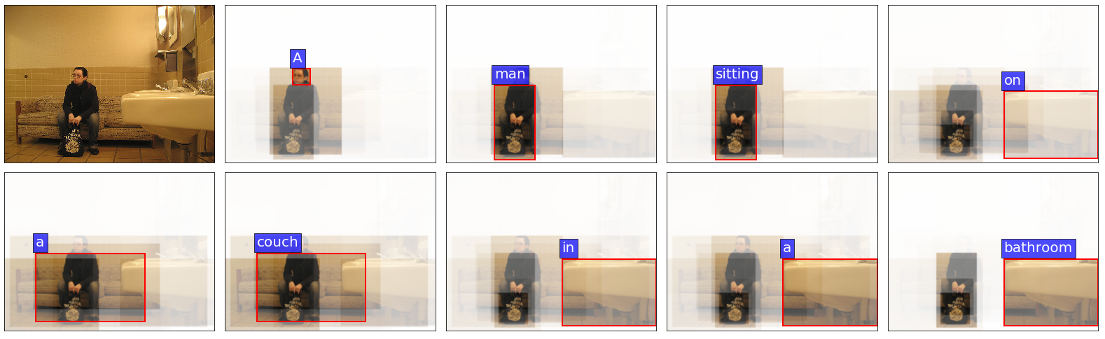}\\
	\end{center}
	\caption{Qualitative differences between attention methodologies in caption generation. For each generated word, we visualize the attended image region, outlining the region with the maximum attention weight in red. The selected image is unusual because it depicts a bathroom containing a couch but no toilet. Nevertheless, our baseline ResNet model (top) hallucinates a toilet, presumably from language priors, and therefore generates a poor quality caption. In contrast, our Up-Down model (bottom) clearly identifies the out-of-context couch, generating a correct caption while also providing more interpretable attention weights. }
	\label{fig:caption_compared}
\end{figure*}

\begin{figure*}[t]
	\begin{center}
		A group of people are playing a video game.	
		\includegraphics[width=1\linewidth]{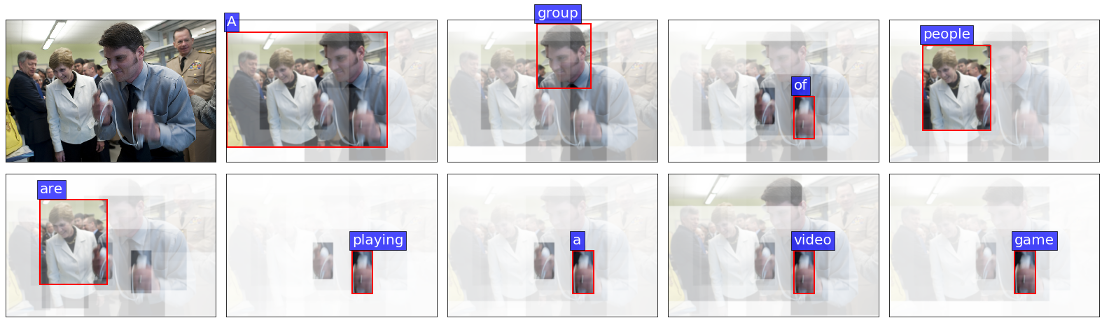}\\
		A brown sheep standing in a field of grass.
		\includegraphics[width=1\linewidth]{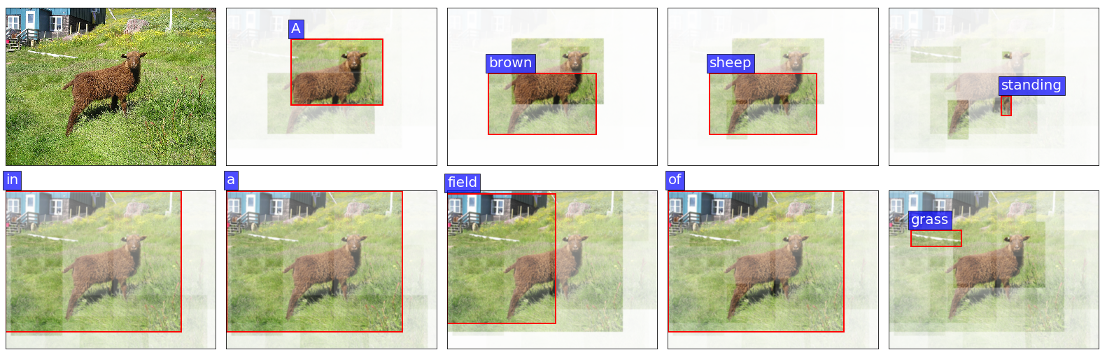}\\
		Two hot dogs on a tray with a drink.
		\includegraphics[width=1\linewidth]{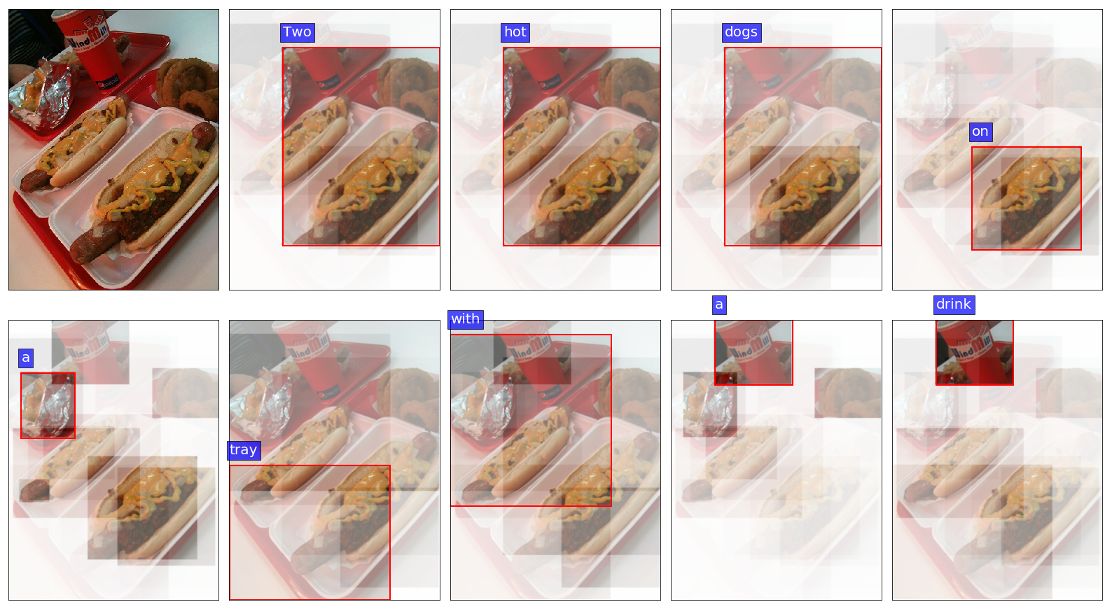}\\
	\end{center}
	\caption{Examples of generated captions showing attended image regions. Attention is given to fine details, such as: (1) the man's hands holding the game controllers in the top image, and (2) the sheep's legs when generating the word `standing' in the middle image. Our approach can avoid the trade-off between coarse and fine levels of detail. }
	\label{fig:caption_examples}
\end{figure*}

\begin{figure*}[t]
	\begin{center}
		Two elephants and a baby elephant walking together.
		\includegraphics[width=1\linewidth]{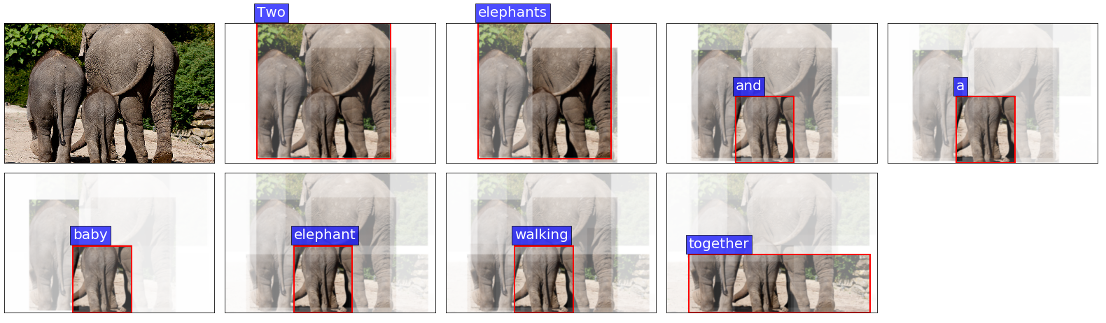}\\
		A close up of a sandwich with a stuffed animal.
		\includegraphics[width=1\linewidth]{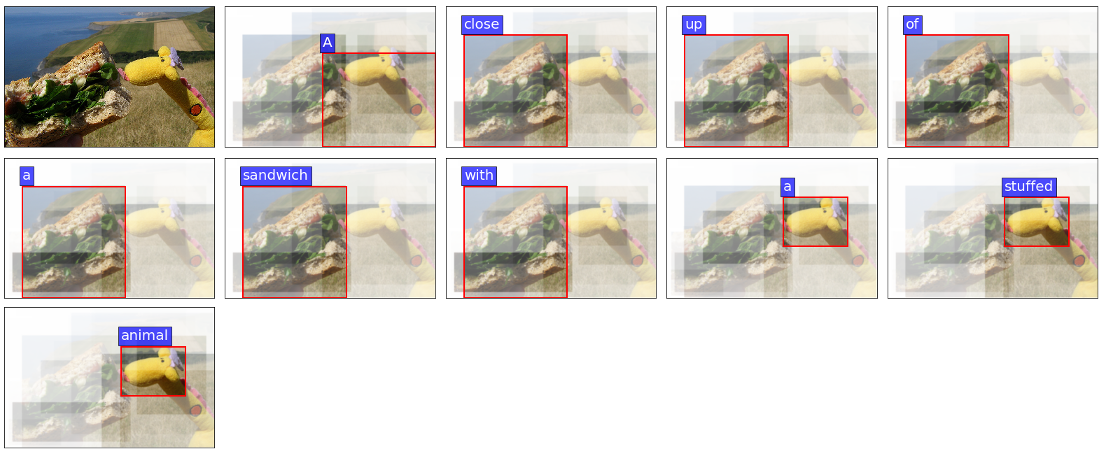}\\
		A dog laying in the grass with a frisbee.
		\includegraphics[width=1\linewidth]{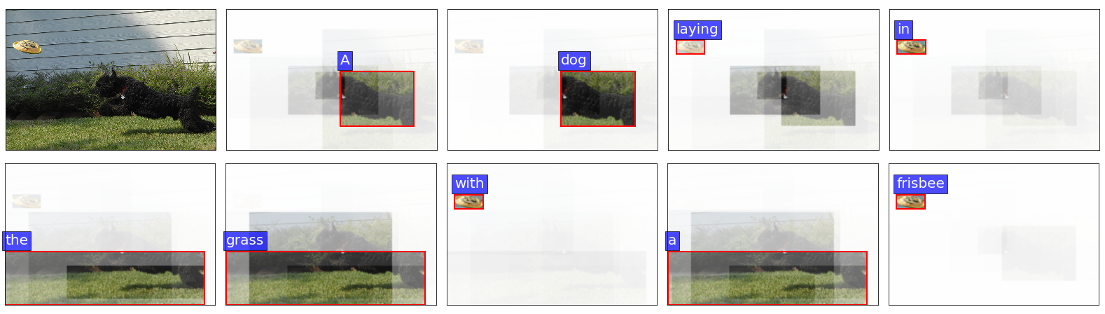}\\
	\end{center}
	\caption{Further examples of generated captions showing attended image regions. The first example suggests an understanding of spatial relationships when generating the word `together'. The middle image demonstrates the successful captioning of a compositionally novel scene. The bottom example is a failure case. The dog's pose is mistaken for laying, rather than jumping -- possibly due to poor salient region cropping that misses the dog's head and feet.}
	\label{fig:extra}
\end{figure*}

\begin{figure*}[t]
	\begin{center}
		Question: What color is illuminated on the traffic light?
		Answer left: green. Answer right: red.
		\includegraphics[width=0.49\linewidth]{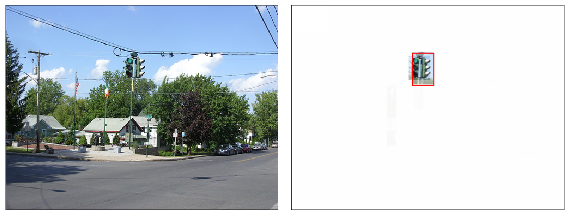}
		\includegraphics[width=0.49\linewidth]{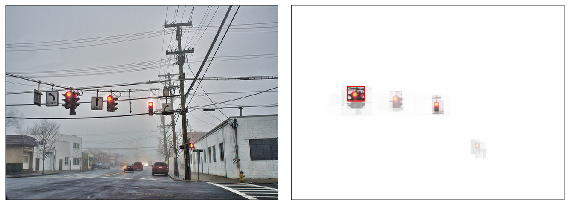}
		Question: What is the man holding?
		Answer left: phone. Answer right: controller.
		\includegraphics[width=0.49\linewidth]{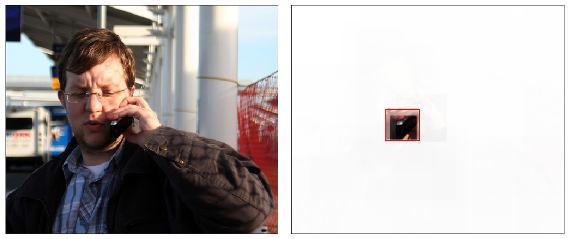}
		\includegraphics[width=0.49\linewidth]{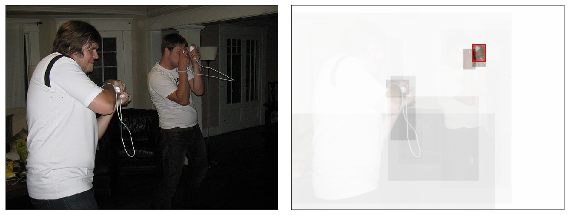}
		Question: What color is his tie?
		Answer left: blue. Answer right: black.
		\includegraphics[width=0.49\linewidth]{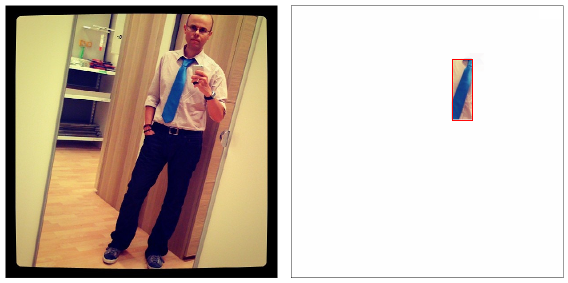}
		\includegraphics[width=0.49\linewidth]{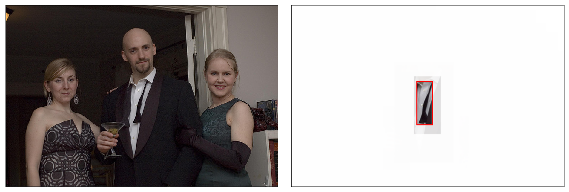}
		Question: What sport is shown?
		Answer left: frisbee. Answer right: skateboarding.
		\includegraphics[width=0.49\linewidth]{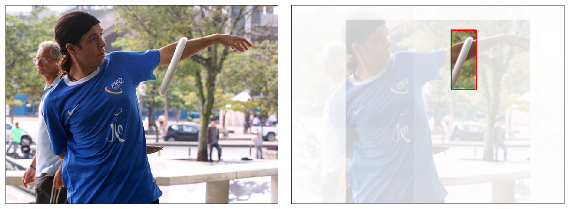}
		\includegraphics[width=0.49\linewidth]{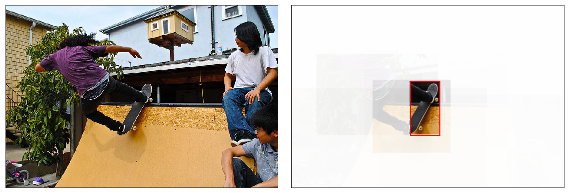}
		Question: Is this the handlebar of a motorcycle?
		Answer left: yes. Answer right: no.
		\includegraphics[width=0.49\linewidth]{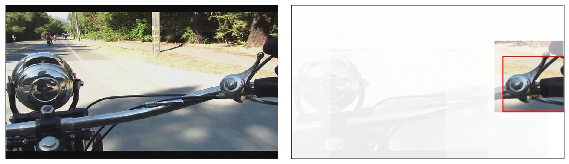}
		\includegraphics[width=0.49\linewidth]{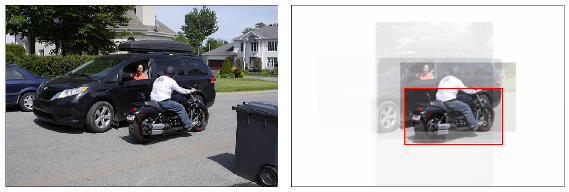}
	\end{center}
	\caption{Further examples of successful visual question answering results, showing attended image regions. }
	\label{fig:vqa_examples}
\end{figure*}

\begin{figure*}[t]
	\begin{center}
		Question: What is the name of the realty company?
		Answer left: none. Answer right: none.
		\includegraphics[width=0.49\linewidth]{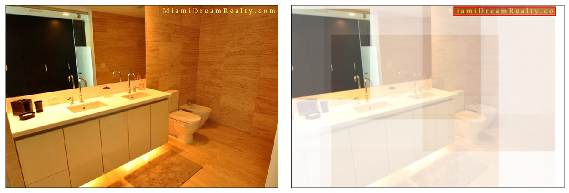}
		\includegraphics[width=0.49\linewidth]{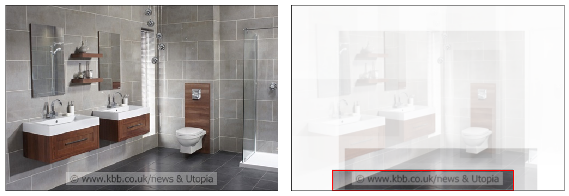}
		Question: What is the bus number?
		Answer left: 2. Answer right: 23.
		\includegraphics[width=0.49\linewidth]{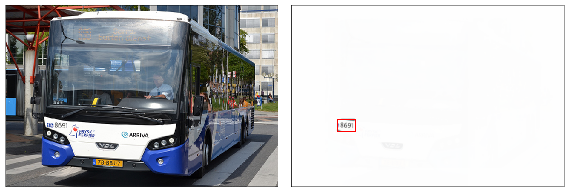}
		\includegraphics[width=0.49\linewidth]{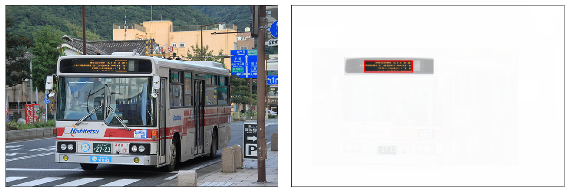}
		Question: How many cones have reflective tape?
		Answer left: 2. Answer right: 1.
		\includegraphics[width=0.49\linewidth]{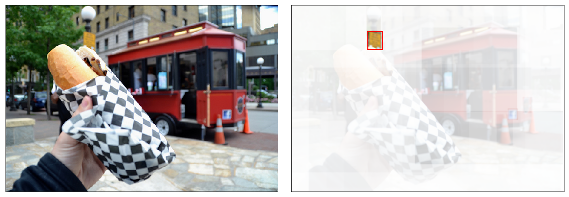}
		\includegraphics[width=0.49\linewidth]{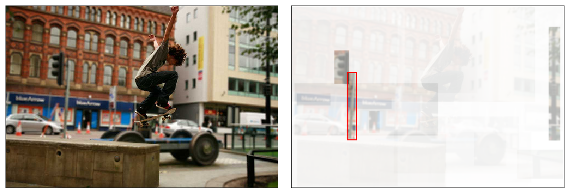}
		Question: How many oranges are on pedestals?
		Answer left: 2. Answer right: 2.
		\includegraphics[width=0.49\linewidth]{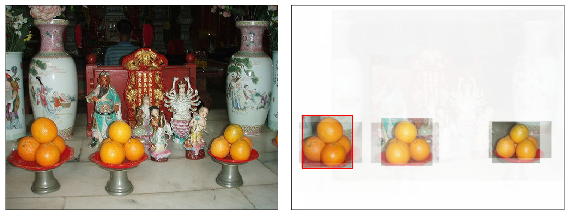}
		\includegraphics[width=0.49\linewidth]{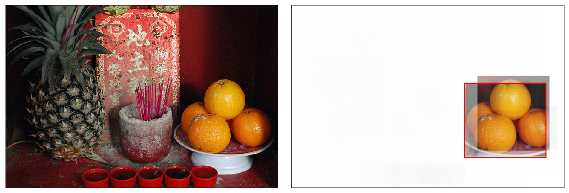}
	\end{center}
	\caption{Examples of visual question answering (VQA) failure cases. Although our simple VQA model has limited reading and counting capabilities, the attention maps are often correctly focused.}
	\label{fig:vqa_fails}
\end{figure*}

In \figref{fig:caption_compared} we qualitatively compare attention methodologies for image caption generation, by illustrating attention weights for the ResNet baseline and our full Up-Down model on the same image. The baseline ResNet model hallucinates a toilet and therefore generates a poor quality caption. In contrast, our Up-Down model correctly identifies the couch, despite the novel scene composition. Additional examples of generated captions can be found in Figures \ref{fig:caption_examples} and \ref{fig:extra}. Additional visual question answering examples can be found in Figures \ref{fig:vqa_examples} and \ref{fig:vqa_fails}.

\end{document}